\documentclass[11pt,a4paper]{article}

\usepackage[margin=1in]{geometry}

\usepackage{times}
\usepackage{latexsym}
\usepackage{booktabs}
\usepackage{amsmath}
\usepackage{amsfonts}
\usepackage{multirow}
\usepackage{enumitem}
\usepackage{siunitx}    
\usepackage{array}      

\usepackage[T1]{fontenc}

\usepackage[utf8]{inputenc}

\usepackage{microtype}

\usepackage{inconsolata}

\usepackage{graphicx}

\usepackage{xcolor}
\usepackage{makecell}
\usepackage[font=small,labelfont=bf]{caption}

\usepackage[round,authoryear]{natbib}
\let\cite\citep
\definecolor{linkblue}{RGB}{25,60,100}
\usepackage{hyperref}
\hypersetup{
  colorlinks=true,
  linkcolor=linkblue,
  citecolor=linkblue,
  urlcolor=linkblue
}

\definecolor{improve}{RGB}{0,115,80}
\definecolor{worsen}{RGB}{175,55,55}

\newcommand{\good}[1]{\textcolor{improve}{#1}}
\newcommand{\bad}[1]{\textcolor{worsen}{#1}}

\usepackage{listings}
\lstdefinestyle{skillpeprompt}{
  basicstyle=\ttfamily\footnotesize,
  columns=fullflexible, keepspaces=true,
  breaklines=true, breakatwhitespace=false,
  showstringspaces=false, frame=single,
  aboveskip=0.6em, belowskip=0.6em
}

\title{\LARGE\bfseries MASIF: Implicit Failures for Process Observability in Multi-Agent Systems}

\author{%
  \begin{tabular}{@{}c@{\qquad}c@{\qquad}c@{}}
    \normalsize Yanwei Huang & \normalsize Arpit Narechania & \normalsize Atanu Sinha \\
    \small HKUST & \small HKUST & \small Adobe Research \\[0.3em]
    \footnotesize\href{mailto:yanwei.huang@connect.ust.hk}{\texttt{yanwei.huang@connect.ust.hk}} &
    \footnotesize\href{mailto:arpit@ust.hk}{\texttt{arpit@ust.hk}} &
    \footnotesize\href{mailto:atr@adobe.com}{\texttt{atr@adobe.com}}
  \end{tabular}%
}
\date{}

\hypersetup{
  pdftitle={MASIF: Implicit Failures for Process Observability in Multi-Agent Systems},
  pdfauthor={Yanwei Huang, Arpit Narechania, Atanu Sinha}
}

\begin{document}
\maketitle
\begin{abstract}

Multi-agent systems (MAS) can complete a task despite local execution errors that are corrected, bypassed, or carried forward without individually determining its outcome. We call these implicit failures (IFs) and study them as signals for process observability beyond final task outcome. Across 1,828 traces from nine MAS--benchmark configurations, IFs occur in 54.3\% of successful and 75.2\% of failed traces. We develop taxonomies of error types and subsequent absorption mechanisms, supported by human and cross-model annotation audits. Counterfactual replay further probes how repairing an IF changes the final outcome. To examine the practical value of IFs, we evaluate IF-based best-of-$N$ trace selection against existing process reward models, and introduce a novel RL algorithm, IF-GRPO, that uses IF costs to refine preferences among successful trajectories in GRPO. Experiments show that IF-GRPO exhibits some gains in accuracy and process metrics. Importantly, it exposes trade-offs between task performance, IF count and cost, and token-normalized IF density. Overall, our results show that IFs provide complementary process observability into how MAS reach their outcomes, including error recovery, propagation, and IF-based process differences that final task success alone does not expose.
\end{abstract}

\section{Introduction}

Large language model (LLM) agents can reason, plan, and interact with environments~\cite{xi2023rise, yao2023react}, but a single agent struggles with complex problems that exceed its context window or require diverse expertise. \emph{Multi-agent systems} (MAS), in which multiple LLM agents collaborate, address this limitation through task decomposition, role-playing, and parallel execution~\cite{tran2025multi, hong2023metagpt, wu2023autogen, li2023camel}.

However, MAS are prone to cascading errors and coordination breakdowns that compromise task success~\cite{cemri2025multi, xie2026spark}, and training them requires step-level rather than trace-level reward signals~\cite{xi2026agentprm}. This has driven a growing line of research on \emph{failure attribution}, which identifies the responsible agent and step in a failed execution trace through causal inference, hierarchical causal graphs, automated diagnosis frameworks, and more realistic benchmarks~\cite{ma2025automatic, wang2026flat, wang2026agenttrace, zhang2025agent, zhang2025agentracer, in2026rethinking, chen2026seeing}.

A key limitation shared by existing approaches is their focus on the \emph{decisive} step that explains an eventual failure. This outcome-oriented view leaves part of the execution process unobserved: a trace may contain local errors that are corrected, bypassed, or carried forward without individually determining its final outcome. We call these \emph{implicit failures} (IFs). IFs can occur in both successful and failed trajectories, revealing how MAS encounter, absorb, and propagate errors during execution. We therefore study IFs as signals for process observability: complementary information about execution behavior that is not captured by the final success label alone.

We study IFs through four research questions:

\textbf{RQ1:} How common are IFs? (Section~\ref{sec:rq1}) Across 1,828 traces from nine MAS--benchmark configurations, IFs occur in 54.3\% of successful and 75.2\% of failed traces, showing that local process errors are prevalent on both sides of the outcome boundary.

 \textbf{RQ2:} Where are IFs from and how are they absorbed? (Section~\ref{sec:rq2}) We develop human-derived taxonomies of IF error types and absorption mechanisms, characterizing how local errors are corrected, bypassed, coordinated around, or propagated, complementing taxonomies that primarily study trace-level failures~\cite{cemri2025multi}.

\textbf{RQ3:} How do MAS outcomes respond to repairing IFs? (Section~\ref{sec:rq3})  Matched counterfactual replay tests whether IFs identify actionable process states. IF repair increases failure rescue by up to 46.2 percentage points over no-op replay.

\textbf{RQ4:} Can IFs provide useful training signals? (Section~\ref{sec:practical}) To further demonstrate IFs' practical utility, we apply them to trace selection and reinforcement learning (RL). Experiments show that our IF-based scoring improves best-of-\(N\) selection over known process reward models (PRMs). We further propose a novel IF-GRPO algorithm (Section~\ref{sec:if_grpo}) that incorporates IFs into Group Relative Policy Optimization, which improves accuracy in six of seven configurations and reduces IF cost per task in three, while exposing trade-offs between task performance and process quality.

Together, these studies position IFs as a complementary layer of process observability for MAS. Rather than asking only whether a system succeeds or which step caused its failure, IFs expose local errors encountered during execution, how the system responds to them, and process-level differences among trajectories with the same outcome.

\section{Methodology} \label{sec:method}

\begin{figure*}[t]
  \centering
  \includegraphics[width=\linewidth]{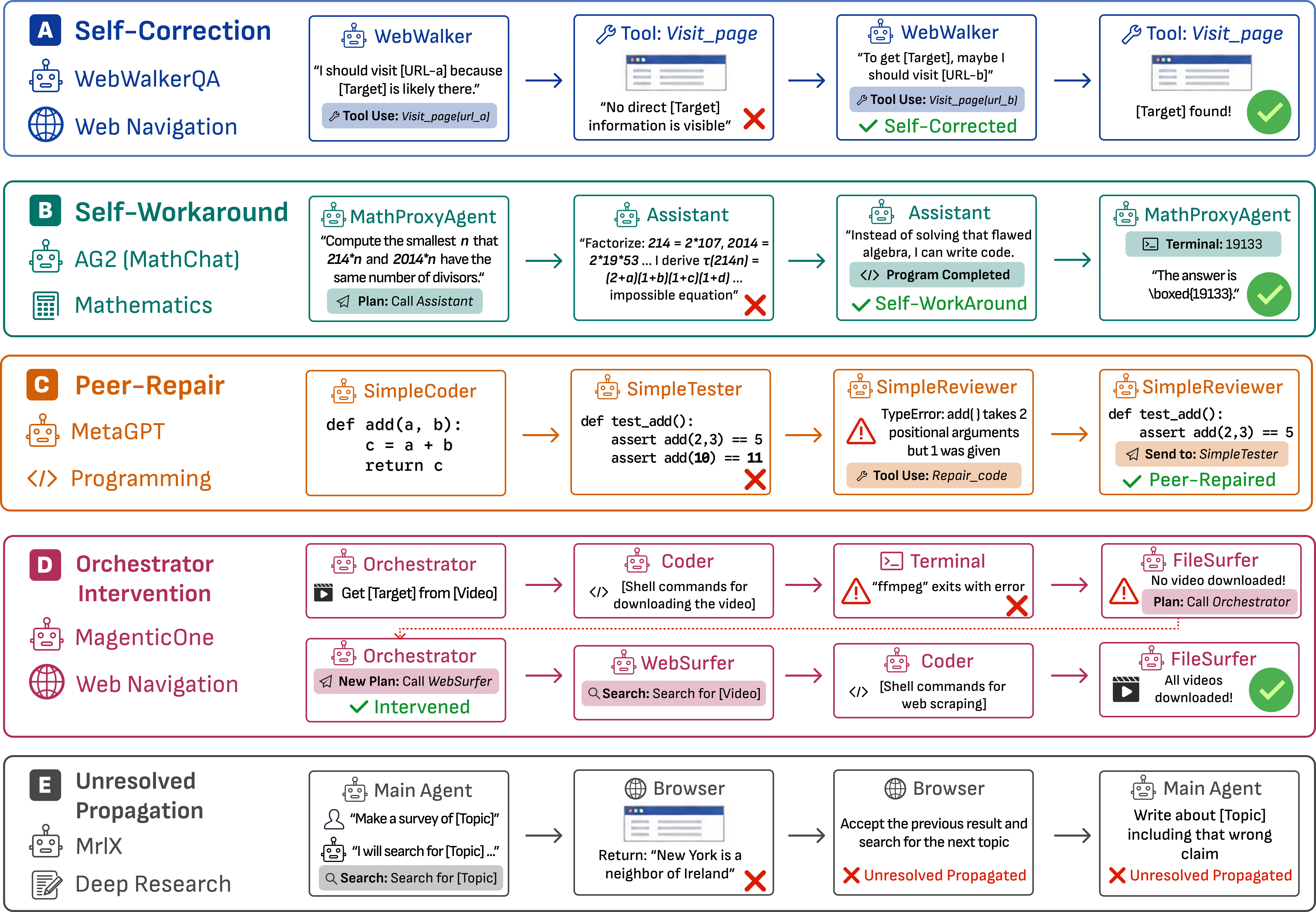}
  \caption{Example MAS traces for five absorption patterns. IFs are marked with crosses, with their resolution steps highlighted. \textit{Self-correction}: WebWalker corrects its own navigation error to find the target. \textit{Self-workaround}: The Assistant switches from flawed algebra to code to solve MathProxyAgent's problem. \textit{Peer-repair}: SimpleReviewer fixes a code error from SimpleTester. \textit{Orchestrator intervention}: The Orchestrator redirects Assistant's failed download to WebSurfer for scraping. \textit{Unresolved propagation}: Browser's wrong claim propagates downstream.}
  \label{fig:absorption}
\end{figure*}

\subsection{Data Collection} \label{sec:collect}
To study the prevalence and patterns of IFs, we constructed a benchmark including 9 MAS-benchmark configurations (Table \ref{tab:datasets}), covering 1,828 MAS traces. We constructed it by selecting 1,048 traces from MAST~\cite{cemri2025multi}, a representative error-annotated benchmark of MAS traces, filtering out traces that lack verifiable ground-truth outcomes or cannot be reproduced.
We further manually curated 780 traces leveraging two popular MAS--(WebWalker~\cite{wu2025webwalker} and MrlX~\cite{mrlx2025}), and benchmarks--(WebWalkerQA~\cite{wu2025webwalker} and BrowseComp-Plus~\cite{chen2025browsecomp}) for web navigation and deep research. 

\begin{table*}[htbp]
\centering
\footnotesize
\setlength{\tabcolsep}{4.5pt}
\caption{Detailed profiles of the configurations in our study corpus.}
\begin{tabular}{lllllll}
\toprule
\textbf{MAS} & \textbf{Benchmark}  & \textbf{Underlying LLM} & \# \textbf{Traces} & \textbf{\# Success} & \textbf{\# Failure} & \textbf{\# IFs} \\
\midrule
Magentic-One & GAIA                 & GPT-4o      & 165 & 52 & 113 & 383 \\
WebAgent (WebWalker)    & WebWalkerQA             & GPT-4o-mini & 680 & 71 & 609 & 1090 \\
MrlX         & BrowseComp-Plus          & Qwen3-8b     & 100 & 4 & 96 & 119 \\
AG2 (MathChat) & OlympiadBench      & GPT-4o & 206 & 29 & 177 & 178 \\
AG2 (MathChat) & GSMPlus &  Claude-3.7-Sonnet \& GPT-4o & 223 & 44 & 179 & 214 \\
AG2 (MathChat) & MMLU     & GPT-4o & 168 & 27 & 141 & 241 \\
ChatDev & MMLU  & GPT-4o & 42 & 23 & 19 & 1 \\
MetaGPT & MMLU  & GPT-4o & 21 & 8 & 13 & 33 \\
HyperAgent & SWE-Bench  & Claude-3.7-Sonnet & 223 & 0 & 223 & 785 \\
\bottomrule
\end{tabular}
\label{tab:datasets}
\end{table*}

Each trace was annotated for IFs using an LLM-as-a-judge pipeline built on
Gemini~3.1 Pro. Given the task context and execution trace,
the judge inspected each step and flagged candidate IFs. 
We validated annotation quality through both human audit and cross-judge analysis. Two annotators independently audited a stratified random sample of 45 traces (5 per configuration). Among 73 existing IF annotations reviewed by both annotators, 64 (87.7\%) were independently retained by both, and none were rejected by both. Separately, another random sample of 90 traces were re-annotated by GLM-5.2, Claude Opus~4.8, and DeepSeek~V4 Pro in addition to the original judge (Gemini 3.1 Pro). Overall, 174 IFs were identified by at least three out of the four judges. Gemini covered 157 (90.2\%) of this set, which were 69.2\% of its identified IFs. 

Annotation completeness is harder to establish. The human audit identified 21 additional IFs across 11 traces, with five independently recovered by both annotators. The missed and disputed cases primarily involve long-range evidence, repeated manifestations of an underlying error, and boundaries between local errors and intermediate task decomposition. Similarly, cross-judge differences concentrate on detection granularity and event boundaries rather than explicit trace-observable errors. A more detailed analysis is provided in Appendix~\ref{sec:reliability}.

\subsection{RQ1: How common are IFs?} \label{sec:rq1}

 RQ1 asks whether IFs are a marginal phenomenon or a pervasive property of MAS
execution. To this end, we measure the fraction of traces containing at least one IF
(\emph{prevalence}) and the mean number of IFs per trace (\emph{intensity}),
stratified by outcome.

As shown in Table \ref{tab:datasets}, across the full corpus, $140$ of $258$ successful traces ($54.3\%$) and
$1{,}181$ of $1{,}570$ failed traces ($75.2\%$) contain at least one IF. Thus, IF presence does not collapse to the final outcome. Over half of successful traces contain IFs, while 24.8\% of failed ones contain none. Therefore, IFs and outcome correctness expose distinct execution properties, motivating a closer examination of what errors occur and how they are absorbed.

\subsection{RQ2: Where are IFs from and how are they absorbed?} \label{sec:rq2}

We investigate the types of local errors underlying IFs and how they are repaired, bypassed, or propagated.

\subsubsection{Study design} To answer this, we had three annotators manually inspect 45 randomly
sampled traces and apply open coding following the grounded theory
~\cite{corbin2017grounded} approach, iterating on the label set for three rounds until it
saturated. The coding process mainly focuses on a) the cause and type of an IF and b) the influence and the eventual outcome of an IF. Afterwards, we followed the same annotation pipeline as described in Section \ref{sec:collect} to label each IF in our benchmark with error types and absorption patterns. The labels were audited by two additional human annotators mentioned in \ref{sec:collect}. Among their jointly retained IFs, agreement was 84.4\% on error type and 89.1\% on absorption pattern.

\begin{table*}[t]
    \centering
    \caption{Taxonomy of IF Types and their share in the benchmark.}
    \label{tab:failure-taxonomy}
    \footnotesize
    \setlength{\tabcolsep}{4.5pt}
    \renewcommand{\arraystretch}{1.3}
    \begin{tabular}{p{2.5cm}p{6cm}p{5cm}p{1cm}}
    \toprule
    \textbf{Type} & \textbf{Description} & \textbf{Example} & \textbf{Share} \\ \midrule
    Failed Tool Use & A tool call fails, is blocked, times out, cannot execute, or returns unusable output because of tool or environment issues. & FileSurfer attempts to read an MP3 audio file as text and receives a conversion exception. & 24.5\% \\ \midrule
    Wrong Action & The agent takes an action that is inappropriate for the current task state or unlikely to make progress. & WebSurfer clicks page text rather than a valid link or play button, leaving the browser on the same page. & 23.0\% \\ \midrule
    Unnecessary Action & The agent performs redundant, low-value, or repetitive actions that consume steps without advancing the task. & WebSurfer handles a cookie prompt instead of following the already identified target page. & 16.4\% \\ \midrule
    Hallucination & The agent fabricates information, tool behavior, source content, or intermediate evidence not supported by the trace. & The agent invents a ranked list item that is not present in the retrieved source. & 15.3\% \\ \midrule
    Incomplete Action & The agent attempts an action but omits required details, formatting, inputs, or follow-through. & FileSurfer repeatedly returns the first PDF page instead of searching or navigating to the relevant pages. & 9.1\% \\ \midrule
    Reasoning Error & The agent makes an incorrect inference, introduces or propagates incorrect information, repeats a known-bad strategy, misses relevant evidence, or terminates from insufficient reasoning. & The assistant swaps document statuses across several listed books, or the orchestrator identifies the wrong single from an album despite available evidence. & 8.6\% \\ \midrule
    Misunderstanding & The agent misunderstands the task, role, constraints, or operational setting. & The orchestrator fails to recognize that another agent has already reached the correct page and restarts the search. & 2.8\% \\ \midrule
    Malformed Output & The agent emits syntactically malformed text or a corrupted structured block, but execution continues. & WebWalker emits a memory block with a stray closing quote and brace before recovering in the next step. & 0.2\% \\ \bottomrule
    \end{tabular}
\end{table*}

\subsubsection{Results} We propose two taxonomies, one regarding the error type and the other regarding the absorption patterns of IFs. The taxonomy of step-level error types is shown in Table \ref{tab:failure-taxonomy}, which is complementary to the trace-level one by Cemri et al.~\cite{cemri2025multi}. Failed tool use (24.5\%) and wrong action (23.0\%) are the two most frequent types, together accounting for 47.5\% of the identified IFs. The taxonomy shares a few similar classes with the one by Cemri et al., such as Wrong/Unnecessary/Incomplete Action (w.r.t. ``Reasoning-Action Mismatch'') and Misunderstanding (w.r.t. ``Disobey Task/Role Specification''). However, our taxonomy provides additional insights into local issues, such as hallucination, tool misuse, and malformed output.

The IF absorption pattern taxonomy consists of six patterns, and we provide a
representative trace for each except  \textit{trivial issue} illustrated in Figure~\ref{fig:absorption}.

\noindent \textbf{Self-correction.} The same agent recognizes and repairs its
    own local error within the same or an immediately subsequent turn. For instance, 
    in Figure~\ref{fig:absorption}A, a WebWalker agent visits the wrong
    education page and, upon receiving a negative result, immediately redirects
    itself to the correct page without external signal.
    
\noindent \textbf{Self-workaround.} 
    The IF is bypassed by the originating agent by advancing the task through a different route, such as choosing a different query, tool, or implementation. In
    Figure~\ref{fig:absorption}B, an AG2 assistant~\cite{AG2_2024} recognizes that its
    algebraic derivation leads to an impossible equation and silently
    substitutes a programmatic search, reaching the correct answer
    without revisiting the flawed reasoning.
    
\noindent \textbf{Peer repair.} The error is compensated for by a different, non-managerial agent downstream. Figure~\ref{fig:absorption}C shows a
    MetaGPT trace in which SimpleTester produces tests that would raise a
    \texttt{TypeError}. SimpleReviewer
    detects the mismatch and issues a corrected version back to SimpleTester.
    
\noindent \textbf{Orchestrator intervention.} A manager or planner peer agent redirects the trajectory after the IF. Figure~\ref{fig:absorption}D, a MagenticOne orchestrator observes that
    local video extraction has failed and replans the entire subtask, switching
    to a WebSurfer-based search strategy that ultimately succeeds.
    
\noindent \textbf{Unresolved propagation.} The IF is left unrepaired with the erroneous state  accepted, carried forward, or folded into later reasoning.  Figure~\ref{fig:absorption}E illustrates an MrlX trace in which
    geographically inconsistent claims returned by the browser are silently
    accepted in the agent's subsequent reasoning.
    
\noindent \textbf{Trivial issue.} A catch-all for absorption patterns with low analytic value, including mechanical retries, duplicate work, ignored or abandoned actions, no-effect errors, lucky coincidences, tool- or environment-level absorption, runtime blocking, timeout, and trace termination.

Table~\ref{tab:absorption} reports the frequency of each absorption label across the corpus, broken down by the outcome of the trace. Several patterns stand out. First, \emph{self-correction} and \emph{self-workaround} are among the most common absorption mechanisms (20.3\% and 14.3\% of all IFs) other than trivial issues, confirming that local recovery within a single agent is the dominant route by which MAS tolerate intermediate errors. Besides, \emph{orchestrator intervention} (9.8\%) and \emph{peer repair} (1.4\%) indicate that inter-agent coordination contributes substantively to IF resolution. We further observe that 17.6\% of IFs are labeled as unresolved propagation. Among the 1,930 IFs with informative absorption labels (excluding the combined trivial/unknown category), 72.2\% are classified as absorbed or repaired and 27.8\% as unresolved propagation. These absorption labels provide process information complementary to final outcome, but are descriptive and do not establish causal effects.

\begin{table}[t]
\centering
\caption{Distribution of IF absorption mechanisms across the corpus, separated by trace outcome.}
\label{tab:absorption}
\small
\setlength{\tabcolsep}{2.3pt}
\begin{tabular}{l r r r r}
\toprule
\textbf{Absorption Mechanism} & \textbf{Success} & \textbf{Failure} & \textbf{Total} & \textbf{Share} \\
\midrule
Self-correction          &  34 & 585 & 619 & 20.3\% \\
Trivial issue            &  77 &    1,037 & 1{,}114 & 36.6\% \\
Orchestrator intervention&  36 &    262 &    298 & 9.8\% \\
Peer repair              &  4 &    38 &    42 &  1.4\% \\
Self-workaround          &  23 &    412 &    435 &  14.3\% \\
Unresolved propagation   &   51 &     485 &     536 &  17.6\% \\
\midrule
\textbf{Total}           & 225 & 2{,}819 & 3{,}044 & 100.0\% \\
\bottomrule
\end{tabular}
\end{table}

\begin{table*}[t]
\centering
\caption{Matched no-op and IF-repair replay across seven
MAS--benchmark configurations.
Arrows indicate no-op $\rightarrow$ IF repair.
SPR measures success preservation and RR measures failure
rescue. $\Delta$RR is the additional rescue rate under IF
repair (percentage points), with 95\% paired task-cluster
bootstrap confidence intervals (5,000 resamples; unadjusted
for multiple comparisons).
Each configuration contains 50 originally successful and
50 failed traces. IF repair increases failure rescue by
23.0--46.2 pp beyond no-op replay in four configurations,
demonstrating the potential of targeted IF interventions
to recover failed executions.}
\label{tab:q3_strict_runtime}

\footnotesize
\setlength{\tabcolsep}{3pt}
\renewcommand{\arraystretch}{1.15}

\begin{tabular*}{\textwidth}{
@{\extracolsep{\fill}}llcccccc@{}
}
\toprule
\textbf{MAS} &
\textbf{Benchmark} &
\textbf{Traces} &
\textbf{Unique Tasks} &
\textbf{IFs} &
\textbf{SPR (\%)} &
\textbf{RR (\%)} &
\textbf{$\boldsymbol{\Delta RR}$ (pp) [95\% CI]} \\
\midrule

AG2 (MathChat) & MMLU
& 100 & 100 & 138
& $100.0\to99.0$
& $1.0\to42.0$
& $+41.0\;[28.6,53.8]$ \\

ChatDev & MMLU
& 100 & 100 & 176
& $96.3\to96.7$
& $8.0\to35.8$
& $+27.8\;[15.8,39.9]$ \\

MetaGPT & MMLU
& 100 & 90 & 145
& $98.0\to91.7$
& $1.0\to47.2$
& $+46.2\;[33.0,60.0]$ \\

\midrule

AG2 (MathChat) & OlympiadBench
& 100 & 98 & 104
& $82.0\to78.0$
& $28.0\to51.0$
& $+23.0\;[7.9,38.4]$ \\

AG2 (MathChat) & GSMPlus
& 100 & 82 & 164
& $89.3\to90.5$
& $42.1\to49.4$
& $+7.3\;[-3.0,17.7]$ \\

MagenticOne & GAIA & 100 & 81 & 206 & $36.7\to40.4$ & $2.7\to7.6$ & $+4.9\;[-2.9, 12.8]$ \\ 

MrlX & BrowseComp-Plus
& 100 & 100 & 209
& $40.2\to39.2$
& $9.5\to5.8$
& $-3.7\;[-11.0,2.5]$ \\

\bottomrule
\end{tabular*}
\end{table*}

\subsection{RQ3: How do MAS outcomes respond to repairing IFs?} \label{sec:rq3}

IFs are locally non-decisive with respect to the observed outcome, but this does not imply that the execution state surrounding an IF is inconsequential. When IFs expose meaningful process state rather than incidental irregularities, locally repairing them may alter subsequent execution. RQ3 tests this possibility through matched counterfactual replay, asking whether IF repair preserves successful executions or rescues failed ones beyond replay-induced variation.

\subsubsection{Preliminaries}
Let $T=(s_1,\ldots,s_n)$ be a MAS trace and
$\mathcal{IF}(T)$ its annotated IF steps.
For each $i\in\mathcal{IF}(T)$, we construct two
continuations from the same execution boundary:
$T_i^{\mathrm{noop}}$, which preserves the original step,
and $T_i^{\mathrm{fix}}$, which replaces it with a local
correction. Let $Y(T)\in\{0,1\}$ denote task correctness.

For either intervention $b\in\{\mathrm{noop},\mathrm{fix}\}$,
we measure success preservation and failure rescue:
\[
\begin{aligned}
\mathrm{SPR}_b(C)
&=P(Y(T_i^b)=1\mid Y(T)=1),\\
\mathrm{RR}_b(C)
&=P(Y(T_i^b)=1\mid Y(T)=0).
\end{aligned}
\label{eq:repair_metrics}
\]
We average over IF positions within each trace and then
over traces in the corresponding outcome stratum.
The incremental rescue effect is
$\Delta\mathrm{RR}=\mathrm{RR}_{\mathrm{fix}}
-\mathrm{RR}_{\mathrm{noop}}$.

\subsubsection{Counterfactual replay}

\textbf{Study Design.} We conduct matched no-op and IF-repair replay across seven configurations in Table~\ref{tab:q3_strict_runtime}. 
Compared with the previously studied configurations (Table ~\ref{tab:datasets}), we excluded HyperAgent because successful traces were difficult to acquire, as well as WebAgent due to environmental issues.
For configurations without sufficient eligible traces on both outcome sides in the original corpus, we regenerated the traces using Qwen3.6-27B with reruns until at least 50 successful and 50 failed traces existed.
Since replaying every trace is prohibitively costly, we sampled 50 successful and 50 failed traces randomly per configuration. The number of unique tasks in the sample is provided in the table. For each selected IF to be repaired, we restored the runtime state at the IF boundary, replaced the IF step with a local correction generated by Gemini 3.1 Pro under the original agent role and context, and resumed MAS execution from the resulting state. The detailed prompt templates are provided in Appendix \ref{app:prompts}.

\textbf{Results.} 
Table~\ref{tab:q3_strict_runtime} shows that IF repair
provides substantial additional failure rescue beyond
no-op replay in four configurations.
The largest gains occur on MetaGPT/MMLU (+46.2 pp)
and MathChat/MMLU (+41.0 pp), followed by
ChatDev/MMLU (+27.8 pp) and
MathChat/OlympiadBench (+23.0 pp) with
their unadjusted 95\% confidence intervals excluding zero.
By contrast, no-op replay already rescues 42.1\% of
failed MathChat/GSMPlus executions, with no clear
additional benefit from repair. 
These findings demonstrate that IF-guided corrections can rescue failed MAS executions beyond replay-induced variation. They highlight the potential of IFs as intervention targets for investigating error recovery.

\section{IFs for MAS Training} \label{sec:practical}
Given the findings, we further ask whether IFs can provide useful training signals in practice (RQ4). We thus test on two downstream settings: inference-time trace selection (Section~\ref{sec:PRM}) and policy optimization (Section~\ref{sec:if_grpo}).

\subsection{Study A: IF as a PRM} \label{sec:PRM}

\textbf{Baselines.} 
We evaluate three settings that cover web navigation and math reasoning: MagenticOne on GAIA~\cite{mialon2023gaia} with PRInTS~\cite{lee2025prints} (165 tasks), and AG2 (MathChat) on GSMPlus~\cite{li2024gsm} with either Qwen2.5-Math-PRM~\cite{zhang2025lessons} or Math-Shepherd~\cite{wang2024math} (80 tasks). 

\textbf{Protocol.} Following the common practice of PRM evaluation, we adopt the best-of-N selection protocol. We divide the tasks into a tuning set and a test set by 20\%-80\%. For each task, we use the MAS to generate $N \in \{4, 8, 16\}$ candidate traces from scratch. We then construct three PRM scorers to generate step-wise scores, which are then aggregated into a trace-level score. The trace with the highest score is selected, and the eventual task accuracy of the selected traces is used to evaluate the scorers. The scorers are introduced below.

\textbf{PRM Scorer.} 
We aggregate step-level PRM scores using four pooling
functions: min, mean, product/geometric mean, and last-step. For each pooling function, we compute a trace\mbox{-}level PRM score for every candidate trace. The candidate with the highest trace\mbox{-}level PRM score is selected, and we compute the resulting task\mbox{-}level accuracy across all tasks in the training set. The pooling function that achieves the highest accuracy in training is selected, held fixed and applied unchanged in testing. Specially, for PRInTS, whose raw step scores range in $[-4, 4]$, we normalize each score to $[0, 1]$ before pooling, while for Qwen2.5-Math-PRM and Math-Shepherd the step scores are already normalized and are directly used as quality scores. The PRM-only selector chooses the highest-scoring candidate among the $N$ traces.

\textbf{IF Scorer.} The IF scorer is a trace\mbox{-}only LLM judge. It reads only the task and the candidate trace without awareness of the expected answer or any oracle information. The judge examines the trace for IFs while noting for each IF its error step, agent, absorption type, and a heuristic severity  designed to assign higher values to unresolved propagated IFs and non-trivial issues. The trace\mbox{-}level IF risk score is calculated by aggregating the severity of all IFs normalized by the trace length. The calculation is detailed in Appendix \ref{sec:severity}, where we further show that candidate rankings are stable to independent \(\pm\)10\% perturbations of these heuristic weights (Spearman \(>0.993\) across 1,000 trials. The IF scorer selects the lowest-risk candidate among the $N$ traces.

\textbf{PRM+IF Scorer.} We further construct a PRM+IF scorer to study the opportunities of incorporating IFs into existing PRMs. To derive the score, we first calculate the scores $s_{prm}, s_{if}$ using the PRM scorer and IF scorer independently. We then perform a weighted average of both scores, i.e., $w \cdot s_{prm} + (1-w) * (1-s_{if})$. Note that the IF scorer's original score $s_{if}$ is transformed into $1-s_{if}$ because the former represents the risk by definition. Similar to the PRM scorer, both the pooling function and the weight $w$ are trained on the training set and then fixed for the testing set.

\begin{table}[t]
\centering
\caption{Best-of-$N$ task accuracy (\%) across 3 configurations (MAS, benchmark, PRM)  and $N \in \{4, 8, 16\}$. IF-based selection outperforms
the evaluated PRMs in all nine comparisons, while
PRM-only selection generally performs close to random,
highlighting the additional process-quality information
captured by IFs.}
\label{tab:prmresult}
\small
\setlength{\tabcolsep}{2.8pt}
\begin{tabular}{p{2.5cm}ccccc}
\toprule
\textbf{Configuration} & \textbf{N} & \textbf{Random} & \textbf{PRM} & \textbf{IF} & \textbf{PRM+IF} \\
\midrule
\multirow{3}{=}{MagenticOne / GAIA / PRInTS} & 4 & 19.3 & 18.2 & \textbf{25.0} & 23.5 \\
 & 8 & 19.8 & 21.2 & \textbf{24.2} & 23.5 \\
 & 16 & 17.7 & 19.7 & \textbf{26.5} & 25.0 \\
\midrule
\multirow{3}{=}{AG2 / GSMPlus / Qwen2.5-Math-PRM} & 4 & 80.1 & 79.7 & \textbf{81.3} & \textbf{81.3} \\
 & 8 & 79.1 & 78.1 & \textbf{82.8} & 81.3 \\
 & 16 & 78.8 & 78.1 & \textbf{82.8} & 78.1 \\
\midrule
\multirow{3}{=}{AG2 / GSMPlus / Math-Shepherd} & 4 & 80.1 & 79.7 & \textbf{81.3} & 79.7 \\
 & 8 & 79.1 & 81.3 & \textbf{82.8} & 81.3 \\
 & 16 & 78.8 & 79.7 & \textbf{82.8} & 79.7 \\
\bottomrule
\end{tabular}
\end{table}

\textbf{Results.} 
Table~\ref{tab:prmresult} shows best-of-$N$ task accuracy across the 
three (MAS, benchmark, PRM) configurations and $N \in \{4, 8, 16\}$. We also present the task accuracy of a random trace selector for reference. Across the
nine settings, PRM-only selection is generally close to random selection. Math-Shepherd and Qwen2.5-Math-PRM, both of which report strong single-agent results in the literature~\cite{wang2024math,zhang2025lessons}, fail to provide a useful selection signal when applied to MAS traces. PRInTS, an agent-trace PRM, performs similarly. By contrast, the IF-only scorer matches or exceeds PRM-only selection in all nine settings, with a gain of 3.0--6.8 percentage points on MagenticOne/GAIA and up to 4.7 points on AG2/GSMPlus. 
The results indicate that IFs provide a useful process-level selection signal across the evaluated MAS and task settings, including settings where the existing PRMs provide little discrimination among candidate traces.

\begin{table*}[t]
\centering
\caption{GRPO versus IF-GRPO across seven MAS--benchmark
configurations. Each configuration reports GRPO $\to$ IF-GRPO
(first row) and its change (second row). ``\# IF'' means the number of IFs.
\good{Green}/\bad{red} indicate improvement/deterioration.
\textbf{Bold} indicates Holm-adjusted $p<0.05$
(two-sided paired permutation tests). IF-GRPO improves accuracy in six
configurations, but per-task IF burden and token-normalized
IF density can change in opposing directions, revealing
process-level trade-offs invisible to task accuracy alone.}
\label{tab:rl_results}

\footnotesize
\setlength{\tabcolsep}{3.8pt}
\renewcommand{\arraystretch}{0.95}

\begin{tabular*}{\textwidth}{
@{}c@{\hspace{10pt}}l@{\extracolsep{\fill}}ccccc@{}
}
\toprule
\textbf{MAS} &
\textbf{Benchmark} &
\textbf{Accuracy (\%)} &
\textbf{IF Cost / Task} &
\textbf{\# IF / Task} &
\textbf{IF Cost / 1K Tokens} &
\textbf{\# IF / 1K Tokens} \\
\midrule

\multirow{6}{*}{\makecell{AG2 \\ (MathChat)}}
& \multirow{2}{*}{MMLU-Pro}
& $54.98\to57.06$
& $1.42\to1.51$
& $1.14\to1.41$
& $4.37\to2.39$
& $3.49\to2.22$ \\[-1pt]
& & \good{+2.08 pp}
& \bad{+6.33\%}
& \bad{+23.84\%}
& \good{\textbf{-45.38\%}}
& \good{\textbf{-36.49\%}} \\[2pt]

& \multirow{2}{*}{GSMPlus}
& $71.75\to72.21$
& $0.49\to0.34$
& $0.49\to0.30$
& $2.33\to1.54$
& $2.31\to1.33$ \\[-1pt]
& & \good{+0.46 pp}
& \good{\textbf{-30.83\%}}
& \good{\textbf{-39.66\%}}
& \good{\textbf{-33.92\%}}
& \good{\textbf{-42.37\%}} \\[2pt]

& \multirow{2}{*}{OlympiadBench}
& $37.69\to40.00$
& $2.35\to2.32$
& $3.37\to2.63$
& $2.40\to1.64$
& $3.45\to1.86$ \\[-1pt]
& & \good{+2.31 pp}
& \good{-0.98\%}
& \good{-21.91\%}
& \good{-31.71\%}
& \good{\textbf{-46.18\%}} \\

\midrule

\multirow{4}{*}{Teamwork}
& \multirow{2}{*}{MMLU-Pro}
& $62.46\to66.20$
& $6.28\to6.36$
& $4.57\to5.46$
& $3.33\to1.29$
& $2.42\to1.11$ \\[-1pt]
& & \good{\textbf{+3.74 pp}}
& \bad{+1.34\%}
& \bad{+19.40\%}
& \good{\textbf{-61.23\%}}
& \good{\textbf{-54.28\%}} \\[2pt]

& \multirow{2}{*}{GSMPlus}
& $77.38\to72.50$
& $1.35\to1.38$
& $1.56\to1.70$
& $1.10\to0.80$
& $1.28\to0.99$ \\[-1pt]
& & \bad{-4.88 pp}
& \bad{+2.30\%}
& \bad{+9.04\%}
& \good{\textbf{-27.50\%}}
& \good{\textbf{-22.77\%}} \\[2pt]

\midrule

\multirow{4}{*}{Self-MoA}
& \multirow{2}{*}{MMLU-Pro}
& $64.70\to65.70$
& $4.39\to4.55$
& $5.06\to5.42$
& $0.92\to0.92$
& $1.06\to1.10$ \\[-1pt]
& & \good{+1.00 pp}
& \bad{+3.69\%}
& \bad{+7.12\%}
& \good{-0.22\%}
& \bad{+3.10\%} \\[2pt]

& \multirow{2}{*}{GSMPlus}
& $75.63\to76.38$
& $0.97\to0.97$
& $1.15\to1.14$
& $0.45\to0.43$
& $0.53\to0.51$ \\[-1pt]
& & \good{+0.75 pp}
& \good{-0.62\%}
& \good{-1.04\%}
& \good{-3.81\%}
& \good{-4.17\%} \\

\bottomrule
\end{tabular*}
\end{table*}

Furthermore, we observe that the PRM+IF scorer performs no better than IF alone. We interpret this as a consequence of 
the previous observation: when the PRM component is itself close to 
random, mixing it with the IF signal adds noise. 

\subsection{Study B: IFs for RL training}
\label{sec:if_grpo}

To further investigate the application of IFs in RL, we incorporate them into Group Relative Policy
Optimization (GRPO)~\cite{shao2024deepseekmath}.

\textbf{GRPO Background.}
For a group of MAS traces, let $y_i\in\{0,1\}$ denote the final
task correctness of trajectory $T_i$. Outcome-only GRPO uses the
group-normalized advantage
\begin{equation}
A_i^{\mathrm{GRPO}} =
\frac{y_i-\bar{y}}{s_y+\epsilon},
\label{eq:out_adv}
\end{equation}
where $\bar{y}$ and $s_y$ denote the group mean and standard deviation.

\textbf{IF-GRPO.}
We introduce IFs as a secondary signal to favor successful traces
with fewer and less severe IFs while preserving task correctness as the
primary objective. For each successful trajectory, an LLM judge
identifies IFs and computes their aggregated cost:
\begin{equation}
C_i = \sum_{e\in\mathrm{IF}(T_i)}
w_{\mathrm{type}}(e)w_{\mathrm{abs}}(e),
\end{equation}
where the weights $w_{\mathrm{type}}, w_{\mathrm{abs}}$ follow Appendix~\ref{sec:severity}.

We apply IF supervision only to \emph{mixed-outcome groups}, which
contain both successful and failed traces. Let
$S=\{i:y_i=1\}$ and $c_{\min},c_{\max}$ denote the minimum and
maximum IF costs among successful traces. We define
\begin{equation}
z_i = \frac{c_{\max}-C_i}{c_{\max}-c_{\min}}-\frac{1}{2},
\quad i\in S,
\label{eq:if_minmax}
\end{equation}
and construct the advantage as
\begin{equation}
A_i =
\begin{cases}
A_i^{\mathrm{GRPO}}(1+z_i), & y_i=1,\\
A_i^{\mathrm{GRPO}}, & y_i=0.
\end{cases}
\label{eq:if_grpo}
\end{equation}
If fewer than two traces succeed or their IF costs are
identical, we set $z_i=0$. All-success and all-failure groups
receive zero advantage. Since $z_i\in[-0.5,0.5]$, IF supervision
adjusts the relative reinforcement of successful traces
without reversing the preference for success over failure.
Thus, IFs refine rather than replace outcome supervision.

\textbf{Setup.}
To examine the applicability of IF-guided optimization across different MAS architectures, we select three representative MAS with distinct coordination mechanisms: AG2 (MathChat), which uses assistant--proxy interaction; Teamwork~\cite{pappu2026multi}, which employs multi-agent discussion; and Self-MoA~\cite{li2025rethinking}, which aggregates independently generated proposals. These systems are evaluated across three benchmarks: MMLU-Pro~\cite{wang2024mmlu}, GSMPlus~\cite{li2024gsm}, and OlympiadBench~\cite{he2024olympiadbench}. All agents use Qwen3-4B as the backbone. Each GRPO group contains eight MAS trajectories. We compare GRPO and IF-GRPO under matched training settings, with Qwen3.8-Flash-Next serving as the IF judge.
Additional details are in Appendix~\ref{sec:additional_imple}.

\textbf{Results.}
As shown in Table~\ref{tab:rl_results}, IF-GRPO yields higher accuracy in six configurations and reduces IF cost
per task in three. Some configurations improve both:
on AG2/GSMPlus, accuracy increases by 0.46 points while
IF cost and count decrease by 30.83\% and 39.66\%.
Others expose trade-offs across outcome and process measures. For example, Teamwork/GSMPlus reduces IF cost and count per 1K tokens significantly while losing 4.88 accuracy points. These divergences show that task accuracy, per-task IF burden, and token-normalized IF density capture different aspects of MAS execution. While IF-GRPO does not establish a uniformly superior training objective, it demonstrates how IF-based process measures expose changes that outcome accuracy alone would conceal.

\section{Related Work}
Despite their effectiveness across domains, MAS are prone to errors, motivating research into their failure patterns. Cemri et al.~\cite{cemri2025multi} analyze 1{,}642 traces across seven MAS frameworks, identifying 14 failure modes spanning specification ambiguity, inter-agent misalignment, and inadequate verification. Wang et al.~\cite{xie2026spark} model error \emph{cascades}, showing how minor errors amplify through recursive context reuse and proposing a governance layer to suppress propagation. Similarly, Reid et al.~\cite{reid2025risk} highlight 6 remarkable failure modes of governed MAS. 

These risks motivate \emph{failure attribution}: identifying the agent and step responsible for a failed trace. Drawing on traditional software debugging, researchers propose tools for prompt testing~\cite{kahng2024llm}, red-teaming~\cite{reid2025risk}, MAS trace visualization~\cite{zhang2025convomap, sheng2026dills}, and trace intervention for what-if testing~\cite{epperson2025interactive}. However, these tools demand substantial developer effort and expertise.

Recent work therefore explores automatic attribution. Zhang et al.~\cite{zhang2025agent} use LLMs to diagnose traces and introduce the Who\&When benchmark with annotations of decisive failures. AgenTracer~\cite{zhang2025agentracer} trains a dedicated attribution model on curated success-failure trajectory pairs, while Ma et al.~\cite{ma2025automatic} use causal inference to predict critical steps without a trained attributor. Wang et al.~\cite{wang2026flat} transform flat execution logs into hierarchical causal graphs for systematic diagnosis, and AgentTrace~\cite{wang2026agenttrace} extends this approach to deployed systems through lightweight post-hoc tracing. DoVer~\cite{ma2025dover} tests automatically generated hypotheses by intervening in MAS traces and measuring outcome flips.

These approaches primarily seek steps that explain an eventual failure. In contrast, we study locally non-decisive errors in both successful and failed traces. IFs complement failure attribution by exposing how errors arise, are absorbed, or propagate during execution, providing process observability beyond the final outcome.

\section{Conclusion}
We present a systematic study of implicit failures (IFs)
in multi-agent systems, analyzing 1,828 traces
across nine MAS--benchmark configurations. We demonstrate
their prevalence 
and characterize their error types and absorption patterns.
Through counterfactual repair, we further show that targeted IF repair can rescue failed executions beyond
the effects of replay alone, suggesting opportunities
for IF-guided debugging and recovery. We also investigate the practical utility of IFs as
process-level supervision. IF-based scoring compares
favorably with established PRMs in best-of-$N$ selection,
while IF-guided GRPO shows both improvements and
trade-offs between task accuracy and IF burden.
Our findings establish IFs as a valuable
dimension of MAS behavior and highlight process-level observability alongside outcome-based
evaluation.

\section*{Limitations}

Our study has several limitations. First, IF identification
relies on LLM-based annotation, which may introduce
judgment bias and miss subtle errors. Although human
audits and cross-judge comparisons support annotation
quality, completeness remains challenging to establish.

Second, while our corpus spans nine MAS--benchmark
configurations, its coverage of task domains, MAS
architectures, and backbone models remains limited.
Our counterfactual replay study examines one IF repair
at a time under a fixed intervention protocol.
Further controlled perturbation experiments could
investigate how MAS respond to injected errors and
multiple simultaneous failures.

Finally, our PRM study focuses on best-of-$N$ selection,
while IF-GRPO training uses a single trainable backbone
with limited training seeds. IF supervision also relies
on a stronger frozen LLM judge, introducing additional
inference costs. Further experiments with diverse
backbones, MAS architectures, and supervision baselines
would help establish the generality and efficiency of
IF-guided optimization.

\bibliography{custom}

\appendix

\section{A Heuristic Algorithm for IF Severity Calculation} \label{sec:severity}

Let an MAS trace contain $m$ IFs and $L$ parsed trace steps. For IF $i$, let
$w_{type}(i)$ be its error-type weight and $w_{abs}(i)$ be its absorption weight. The trace-level IF risk $r$ is:

\[r = \frac{\sum_i w_{type}(i) * w_{abs}(i)}{max(1, L)}\]

The score is therefore higher when a trace contains more IFs, more intrinsically
dangerous IF types, less effective absorption mechanisms, or a higher density
of IFs per trace step. 

We assign heuristic values to $w_{type}$ and $w_{abs}$, as illustrated in Table \ref{tab:wtype} and Table \ref{tab:wabs}. For $w_{type}$, the highest weights are assigned to Hallucination and Reasoning Error since fabricated evidence and incorrect inferences silently contaminate downstream reasoning and directly degrade answer quality. Mid-range weights cover Wrong Action and Misunderstanding as bad navigation, tool, or planning actions can derail a trajectory but are often recoverable, while misreading task or role constraints is serious because subsequent steps may optimize for the wrong objective. Lower weights apply to Failed Tool Use, Malformed Output, Incomplete Action, and Unnecessary Action. Tool failures are frequently externally caused and retryable, missing follow-through is often repaired by the same or another agent, and redundant work mostly consumes budget without inflicting semantic damage. 

For $w_{abs}$, the lowest weight is reserved for Trivial Issue, capturing failures that are mechanical, redundant, ignored, or otherwise absorbed without substantive consequences. Self-correction sits slightly higher, since the originating agent itself detects and repairs the local failure, while Self-workaround reflects the same agent bypassing the issue via a different route rather than directly fixing it. Mid-range weights describe repairs that depend on other components: Peer Repair captures useful compensation by another agent at the cost of added coordination dependence, and Orchestrator Intervention reflects a manager or planner redirecting the trajectory while leaving the original worker state unrepaired. The highest weight, Unresolved Propagation, signals that the IF was carried forward or implicitly accepted, leaving the local failure active downstream and fully contributing to residual risk.

While our current weights are derived from qualitative analysis of error severity, we conducted a sensitivity study. On best-of-\(N\) selection, we perform 1,000 random perturbation trials, where each weight was independently varied by up to $\pm 10 \%$. We found that the resulting candidate rankings remained highly consistent with the original rankings, with the global Spearman rank correlation remaining above 0.993 across trials. On RL training, we vary the IF type and absorption weights on AG2/GSMPlus
while keeping the remaining training protocol fixed.
Table~\ref{tab:weight_sensitivity} reports the results. Our original values achieve the lowest per-task IF cost and count among the evaluated weight settings, while being suboptimal on IF count density. However, it also exhibits limitations on IF cost density. We leave a more systematic exploration of optimal  parameters as future work. 

\begin{table}[h]
\caption{The heuristic values of $w_{type}$ for each error type.}
\label{tab:wtype}
\centering
\begin{tabular}{lr}
\toprule
\textbf{Error Type} & \textbf{$w_{type}$} \\
\midrule
Hallucination & 1.30 \\
Reasoning Error & 1.20 \\
Wrong Action & 1.10 \\
Misunderstanding & 1.10 \\
Failed Tool Use & 1.00 \\
Malformed Output & 1.00 \\
Incomplete Action & 0.90 \\
Unnecessary Action & 0.60 \\
\bottomrule
\end{tabular}
\end{table}

\begin{table}[h]
\caption{The heuristic values of $w_{abs}$ for each absorption pattern.}
\label{tab:wabs}
\centering
\begin{tabular}{lr}
\toprule
\textbf{Absorption Pattern}  & \textbf{$w_{abs}$} \\
\midrule
Trivial Issue  & 0.35 \\
Self-correction & 0.55 \\
Self-workaround & 0.65 \\
Peer Repair & 0.75 \\
Orchestrator Intervention & 0.80 \\
Unresolved Propagation & 1.40 \\
\bottomrule
\end{tabular}
\end{table}

\begin{table*}[t]
\centering
\caption{Sensitivity to IF severity weights on AG2/GSMPlus.
Uniform sets the corresponding weights to one.
Perturbed changes the original weights by up to 10\%.
All metrics are measured on the same evaluation benchmark.}
\label{tab:weight_sensitivity}
\setlength{\tabcolsep}{3pt}
\begin{tabular}{lccccc}
\toprule
\textbf{Weights} & \textbf{Accuracy}
& \textbf{IF Cost / Task} & \textbf{\# IF / Task}
& \textbf{IF Cost / 1K Tokens} & \textbf{\# IF / 1K Tokens} \\
\midrule
Original           & 72.21 & \textbf{0.34} & \textbf{0.30} & 1.54 & 1.33  \\
Uniform absorption & 72.42 & 0.36 & 0.45 & 0.84 & \textbf{1.04} \\
Uniform type       & 71.96 & 0.57 & 0.95 & 1.07 & 1.79 \\
Perturbed          & \textbf{73.29} & 0.46 & 0.60 & \textbf{1.04} & 1.37 \\
\bottomrule
\end{tabular}
\end{table*}

\section{Data Consent and Ethical Considerations} \label{sec:consent}

Our study involved a small number of human annotators for 
two purposes: (a) auditing IF annotations produced by the LLM judge, and 
(b) performing open coding to derive the taxonomies described in 
Section~\ref{sec:rq2}. All annotators were members of the research team 
who participated voluntarily as part of their research role. Prior to commencing their tasks, each participant was thoroughly briefed on the research objectives, the specific use of their contributions (validating automated IF tags and building the taxonomies presented herein), the commitment to aggregate and report all data anonymously, and their unrestricted right to opt out at any stage. Formal informed consent was secured from every annotator before the evaluation phase began. In accordance with privacy standards, this study neither gathered nor published any demographic details, personal identifiers, or sensitive information regarding the participants. Additionally, the research protocol was formally reviewed and authorized by our institution's Human Research Ethics Committee.

\subsection{Open Coding Instructions for Annotators}
The goal of this round is to explore and map the space of implicit failure-related phenomena in MAS traces.
All coders will read a small sample of MAS traces from a benchmark we curated.
Use open coding to mark segments that are relevant to MAS implicit failures (broadly construed).
Generate initial codes (short labels + notes) describing what is happening at those segments.
Bring these codes to later meetings, where they will be clustered into higher-level concepts and tentative failure types.
There is no expectation that you agree perfectly in this round. Diversity in codes is valuable.

For each assigned trace:
Skim the whole trace first
Get a sense of the task, involved agents, and how the trace ends (success/failure/mixed).
Note roughly where things start to go wrong (if they do).
Hint: The traces are lengthy with much repetitive info. You may use a LLM to help with your interpretation.
Do line-by-line / segment-by-segment reading in the interface
The interface will show the trace as a sequence of steps.
Each ``step'' may be one agent message, one tool call + response, or a small contiguous block of log lines.
Mark segments that seem relevant to implicit failure.
Highlight the segment (a step, or a portion of a step).
Attach one or more open codes and a brief note (1--3 sentences).
Repeat until you reach the end of the trace.
Not every step needs a code.
After each trace, spend 5--10 minutes reviewing your own codes
Merge obviously duplicated codes.
Clarify unclear labels.
Jot down any emerging patterns you notice (for discussion later).

\section{Auditing IF Annotation}
  \label{sec:reliability}

 \textbf{Previously unidentified IFs.}
 The human audit identified 21 previously unlabeled IFs across 11 traces. The authors of the paper inspected these IFs and found that they concentrated in several recurring patterns: A prominent group involved unsupported execution-state claims in long software-engineering traces. For example, an agent reported that tests had passed without executing them, or proceeded as though a code modification had been applied when the underlying codebase had not changed. Other omissions involved incomplete actions that became visible only through a later orchestrator response, such as opening a date filter without setting the requested date, or navigating to a search service without issuing the requested query. We also observed malformed tool arguments and intermediate outputs that were subsequently consumed by later steps.

  The five IFs independently recovered by both annotators provide the clearest examples. Two occurred in a Magentic-One trace in which a worker repeatedly failed to apply the requested date filter and was redirected by the orchestrator. In an AG2/GSMPlus trace, the solver introduced an unsupported numerical assumption that was used in later reasoning, although the system still reached the correct answer. In a WebWalker trace, the agent searched using a year that contradicted the task requirement; in another, an HTML fragment was accidentally included in a button argument, producing an invalid tool action.

  Cases identified by only one annotator were concentrated in situations requiring more global interpretation. Long HyperAgent traces, for example, require the reviewer to connect a claimed code or test state with execution evidence appearing much earlier in the trace. Other cases lie near the boundary between an incomplete action and ordinary task decomposition: one annotator may regard a worker's partial response as an error, while another may regard it as an intermediate result that the orchestrator is expected to refine. Persistent errors also introduce granularity differences, since a reviewer may annotate each repeated manifestation while another records only the earliest source event. These cases explain why independently discovered omissions exhibit lower agreement than review of an already localized IF.

  \textbf{Cross-judge agreement and disagreement.}
  The four-LLM-judge audit produced 174 events supported by at least three judges, including 81 detected by all four. Agreement was strongest for events with explicit trace evidence, such as malformed tool arguments, unsupported intermediate claims, failed actions followed by an observable repair, and incomplete operations explicitly recognized by an orchestrator.

  The main cross-judge difference was the detection boundary rather than failure to recognize a common core. Claude and GLM tended to annotate more fine-grained or repeated process irregularities, whereas Gemini and DeepSeek applied a more selective threshold. Judges also differed in whether repeated manifestations of the same underlying mistake should be represented as separate IFs. This pattern helps explain how Gemini can cover 90.2\% of the majority-supported events while some events detected by only one or two judges remain outside its annotation set.

  \textbf{Directions for improving annotation.}
  Our current selection of Gemini 3.1 Pro to identify IFs is based on a rationale where precision is prioritized to recall. Our audits suggest several practical improvements that we plan to implement in the future work. First, persistent manifestations of the same underlying error should be grouped under a common event identifier, with the earliest source step distinguished from later propagation steps. Second, long or nested traces would benefit from explicit links between claims about tool execution, code modification, or test results and the corresponding execution evidence. Finally, multi-judge agreement can be retained as an uncertainty signal rather than forcing every candidate into an equally confident binary label. 

\section{Additional Implementation Details} \label{sec:additional_imple}

\subsection{Details of the GRPO and IF-GRPO studies}

We provide the additional implementation details of the GRPO and IF-GRPO studies (Section \ref{sec:if_grpo}) below.

\textbf{Data splits.}
AG2/GSMPlus uses 650 GSM8K training questions, 130 GSM8K validation
questions, and 2,400 GSMPlus questions for testing.
AG2/OlympiadBench uses fixed partitions of 650/130/130 questions
for training/validation/testing. 
Validation questions are excluded from
policy updates.

\textbf{Training parameters.}
We use AdamW with constant learning rate $10^{-6}$,
$(\beta_1,\beta_2)=(0.9,0.98)$, weight decay 0.1, and PPO ratio-clipping
bounds $[0.8,1.28]$. Both the reference-policy KL coefficient and entropy
regularization coefficient are zero. Training decoding uses temperature
0.7 and top-$p=0.95$, with separate thinking mode disabled, while evaluation
uses greedy decoding. AG2 allows up to
15 assistant calls, 2,048 generated tokens per call, and 8,192 per trace.
Teamwork uses 21 calls (four initial answers, four discussion rounds,
and one final answer), with 1,024 tokens per call. 

\textbf{IF scoring and evaluation.}
In IF-GRPO, IF cost only modulates positive
outcome advantages for successful traces in mixed groups. Failed-trace
advantages are unchanged, and all-success/all-failure groups have zero
advantage. Entire zero-advantage batches skip optimizer updates.
Outcome advantages use sample-standard-deviation normalization with
$\epsilon=10^{-6}$. Qwen3.8-Flash-Next uses temperature 0 for IF detection
and absorption labeling.

\subsection{Prompt templates} \label{app:prompts}
We provide the prompt templates we use for IF detection (Listings \ref{lst:if-detect-system} and  \ref{lst:if-detect-user}), absorption pattern annotation (Listings \ref{lst:if-abs-system} and \ref{lst:if-abs-user}), and counterfactual replay (Listings \ref{lst:repair-system} and  \ref{lst:repair-user-blind}) below. 
\begin{lstlisting}[breaklines=true,frame=single,basicstyle=\ttfamily\small,style=skillpeprompt,caption={The system prompt template for IF detection.},label={lst:if-detect-system}]
You are an expert analyzer of LLM-based multi-agent system execution traces.

Identify implicit failures using only the task prompt and execution trace. An implicit failure is a concrete, trace-observable process error that does not individually determine the final outcome of the trace. Determine whether the event meets this definition, but do not classify its absorption pattern.

Return only valid JSON matching the requested schema, with no markdown or extra text.
\end{lstlisting}

\begin{lstlisting}[breaklines=true,frame=single,basicstyle=\ttfamily\small,style=skillpeprompt,caption={The user prompt template for IF detection.},label={lst:if-detect-user}]
Analyze this multi-agent execution trace for concrete local process errors.

## Task
{{TASK_PROMPT}}

## Execution Trace
{{EXECUTION_TRACE}}

## Instructions

Identify all clear local implicit errors visible in the trace. An implicit error is a specific mistake, failed action, unsupported claim, malformed tool use, invalid inference, misunderstanding, omission, hallucination, or materially unnecessary action at a particular trace step, which does NOT individually determine the final outcome of the trace.

For each implicit error, return:
- error_step: the step index at which the error occurs
- error_agent: the agent responsible for the error
- error_type: one of failed_tool_use, wrong_action, unnecessary_action,  hallucination, incomplete_action, reasoning_error, misunderstanding, malformed_output
- error_description: a concise description of what is wrong, grounded in trace evidence
- error_evidence: a short excerpt or close paraphrase of the trace evidence
- severity: low, medium, or high, based only on the local process error
- confidence: high, medium, or low

Detailed explanation of the error type taxonomy:

| Type | Description | Example |
| --- | --- | --- |
| Failed Tool Use | A tool call fails, is blocked, times out, cannot execute, or returns unusable output because of tool or environment issues. | FileSurfer attempts to read an MP3 audio file as text and receives a conversion exception. |
| Wrong Action | The agent takes an action that is inappropriate for the current task state or unlikely to make progress. | WebSurfer clicks page text rather than a valid link or play button, leaving the browser on the same page. |
| Unnecessary Action | The agent performs redundant, low-value, or repetitive actions that consume steps without advancing the task. | WebSurfer handles a cookie prompt instead of following the already identified target page. |
| Hallucination | The agent fabricates information, tool behavior, source content, or intermediate evidence not supported by the trace. | The agent invents a ranked list item that is not present in the retrieved source. |
| Incomplete Action | The agent attempts an action but omits required details, formatting, inputs, or follow-through. | FileSurfer repeatedly returns the first PDF page instead of searching or navigating to the relevant pages. |
| Reasoning Error | The agent makes an incorrect inference, introduces or propagates incorrect information, repeats a known-bad strategy, misses relevant evidence, or terminates from insufficient reasoning. | The assistant swaps document statuses across several listed books, or the orchestrator identifies the wrong single from an album despite available evidence. |
| Misunderstanding | The agent misunderstands the task, role, constraints, or operational setting. | The orchestrator fails to recognize that another agent has already reached the correct page and restarts the search. |
| Malformed Output | The agent emits syntactically malformed text or a corrupted structured block, but execution continues. | WebWalker emits a memory block with a stray closing quote and brace before recovering in the next step. |

Return this JSON schema:
{
  "implicit_failures": [
    {
      "error_step": 5,
      "error_agent": "AgentName",
      "error_type": "reasoning_error",
      "error_description": "The agent reports an unsupported and internally inconsistent count.",
      "error_evidence": "The relevant trace evidence.",
      "severity": "medium",
      "confidence": "high"
    }
  ],
  "summary": {
    "total_implicit_failures": 1,
    "overall_process_assessment": "Brief trace-only assessment."
  }
}

If no clear local errors are visible, return an empty implicit_failures list.
\end{lstlisting}

\begin{lstlisting}[breaklines=true,frame=single,basicstyle=\ttfamily\small,style=skillpeprompt,caption={The system prompt template for IF absorption pattern annotation.},label={lst:if-abs-system}]
You are an expert analyzer of how multi-agent systems handle local process errors.

Classify the absorption pattern using only the task prompt, the given local-error annotation, and execution-trace evidence. Do not use or infer the expected answer, final-answer correctness, trace outcome, success/failure split, or decisive-error annotation. Do not assume that a correct final answer makes an error trivial.

Return only valid JSON matching the requested schema, with no markdown or extra text.
\end{lstlisting}

\begin{lstlisting}[breaklines=true,frame=single,basicstyle=\ttfamily\small,style=skillpeprompt,caption={The user prompt template for IF absorption pattern annotation.},label={lst:if-abs-user}]
Classify how the execution trace handles the annotated local error.

## Task
{{TASK_PROMPT}}

## Local error annotation
{{ERROR_ANNOTATION}}

## Trace context
{{EXECUTION_TRACE}}

## Absorption labels

- self_corrected: The same agent recognizes and repairs its own local error.
- self_workaround: The originating agent bypasses the local error through a genuinely different route, query, tool, source, or implementation strategy.
- peer_repair: A different non-managerial agent corrects the error or supplies a replacement result that is used downstream.
- orchestrator_intervention: A manager, planner, or orchestrator recognizes the problem and redirects or corrects the trajectory.
- unresolved_propagation: The local error is not repaired before its erroneous artifact is accepted or used by at least one downstream step as evidence, a premise, memory, goal, calculation input, action/tool input, patch input, or justification.
- trivial_issue: A catch-all for absorption patterns with low analytic value, including mechanical retries that do not change strategy, duplicate work, ignored or abandoned actions, tool- or environment-level absorption, runtime blocking, timeout, and trace termination. Also use it when the available trace evidence is insufficient to distinguish a trivial issue from another absorption pattern.

## Clarification
- To distinguish between self_corrected, self_workaround, and trivial_issue, the decisive dimension is how the originating agent proceeds. Apply self_corrected when the same agent repairs the faulty output while keeping the same core tool, query, implementation logic, or data source--for example, by adjusting meaningful input parameters or recalculating within the same reasoning chain. Apply self_workaround only when the agent abandons the original failing method and instead pursues a genuinely different route--such as switching to an alternative tool, querying a different data source, or adopting a different implementation path. An exact or merely mechanical retry, including a superficial formatting-only retry, belongs to trivial_issue.
- To distinguish between peer_repair and orchestrator_intervention, the decisive dimension is the hierarchical role of the corrective agent and the scope of the correction. Apply peer_repair when a non-managerial agent (i.e., at the same or equivalent execution level as the original agent) provides a corrected output or replacement result, and this correction is integrated locally without triggering a global re-plan or trajectory redirection. Apply orchestrator_intervention only when a manager, planner, or orchestrator explicitly detects the error and imposes a high-level change--such as issuing a revised plan, reassigning the subtask to a different agent, aborting and restarting the workflow with new instructions, or actively overriding the erroneous step as part of a systemic trajectory adjustment. In practice, look for evidence of top-down command or plan modification for orchestrator_intervention; if the fix comes from a same-level collaborator without altering the overall task topology, use peer_repair.
- To distinguish between unresolved_propagation and trivial_issue, the decisive factor is whether the erroneous artifact--the faulty output, incorrect calculation, or flawed premise--is demonstrably consumed by at least one downstream step. Apply unresolved_propagation only when the trace provides clear evidence that the artifact is explicitly used as an input, evidence, premise, memory update, or justification in a subsequent action, regardless of whether the overall task ultimately succeeds. Conversely, apply trivial_issue if the issue is contained locally by a mechanical retry, ignored, or occurs in a terminated or abandoned branch and does not demonstrably feed into later processing.

Choose exactly one label. Return:
{
  "absorption_pattern": "self_corrected|self_workaround|peer_repair|
orchestrator_intervention|unresolved_propagation|
trivial_issue",
  "confidence": "high|medium|low",
  "rationale": "One concise sentence grounded in the downstream trace evidence."
}
\end{lstlisting}

\begin{lstlisting}[breaklines=true,frame=single,basicstyle=\ttfamily\small,style=skillpeprompt,caption={The system prompt template for counterfactual local repair.},label={lst:repair-system}]
You are an expert in LLM-based Multi-Agent Systems. Given a partial execution trace where a specific agent step was wrong, you produce a corrected replacement of that single step in the agent's voice. You must:

- Preserve the agent's persona, tool-call format, and writing style as evident from the trace.
- Fix only the identified error; do not anticipate or pre-empt later steps.
- Output ONLY the corrected message body. No commentary, no JSON wrapping, no leading "FixedMessage:" labels. Just the literal text the agent should have produced.

\end{lstlisting}

\begin{lstlisting}[breaklines=true,frame=single,basicstyle=\ttfamily\small,style=skillpeprompt,caption={The user prompt template for counterfactual local repair.},label={lst:repair-user-blind}]
## Task being solved by the multi-agent system
{question}

## Conversation history up to (but NOT including) the buggy step

{prefix}

## Buggy step (step {step_index}, agent: {agent}) -- this is what the agent actually said and what we want to replace
<<<BUGGY MESSAGE BEGIN>>>
{buggy_message}
<<<BUGGY MESSAGE END>>>

## What was wrong with this step (annotation from an analyzer)
- error_type: {error_type}
- description: {error_description}
- severity: {severity}

## Your job
Produce a single corrected version of the buggy step that the {agent} agent should have output instead.
Stay in the same persona and format. Do not include any text other than the message body itself.
\end{lstlisting}

\end{document}